# Latent Factorization of Tensors with Threshold Distance Weighted Loss for Traffic Data Estimation

*Lei Yang[1]*

[1] *Chongqing University of Posts and Telecommunications, Chongqing, China*

**Abstract**

Intelligent transportation systems (ITS) rely heavily on complete and high-quality spatiotemporal traffic data to achieve optimal performance. Nevertheless, in real-word traffic data collection processes, issues such as communication failures and sensor malfunctions often lead to incomplete or corrupted datasets, thereby posing significant challenges to the advancement of ITS. Among various methods for imputing missing spatiotemporal traffic data, the latent factorization of tensors (LFT) model has emerged as a widely adopted and effective solution. However, conventional LFT models typically employ the standard $L2$-norm in their learning objective, which makes them vulnerable to the influence of outliers. To overcome this limitation, this paper proposes a threshold distance weighted (TDW) loss-incorporated Latent Factorization of Tensors (TDWLFT) model. The proposed loss function effectively reduces the model's sensitivity to outliers by assigning differentiated weights to individual samples. Extensive experiments conducted on two traffic speed datasets sourced from diverse urban environments confirm that the proposed TDWLFT model consistently outperforms state-of-the-art approaches in terms of both in both prediction accuracy and computational efficiency.

## 1  Introduction

As artificial intelligence continues to advance, intelligent transportation systems(ITS) have become increasingly vital to the construction of smart cities. By integrating cutting-edge technologies such as big data analytics, cloud platforms, and mobile internet, ITS facilitates real-time traffic monitoring, data-driven analysis, and dynamic control [1]. These capabilities contribute significantly to enhancing travel efficiency and safety [2].
Traffic data, encompassing various types of information, including traffic flow, speed, and vehicle trajectories, serves as a fundamental component of ITS[3],[4]. These data are primarily gathered through two types of sensors: fixed sensors and mobile sensors. Since complex sensor faults, unstable communication links, and inadequate sensor deployment, and extreme weather conditions, the resulting traffic data are often incomplete[5],[6] and contain outliers [7], [8],[9],[10]. Therefore, the accurately and efficiently recovery of incomplete spatiotemporal traffic data has become a crucial and challenging research problem.

To date, researchers have proposed various solutions to address the aforementioned issues, among which latent factorization of tensors(LFT)[11, 12],[13],[14] have emerged as promising solution due to its notable scalability and accuracy[15],[16]. Specifically, a LFT-based models typically represent temporally dynamic traffic data using a high-dimensional and partially observed third-order tensor structured as "road ×day ×interval". Such a representation supports the precise modeling of the intricate spatio-temporal correlations[17] characteristic of traffic systems. Unlike traditional tensor decomposition approaches that operate on complete tensors, LFT frameworks define their objective function and parameter learning procedures solely on the observed data[18],[19],[20, 21],[22],[23] with missing data excluded[24]. Therefore, designing a reliable LFT-based imputation model requires careful formulation of both the optimization objective and the learning mechanism tailored to incomplete observations.
According to prior studies[25], [26],[27],[28],most LFT models formulate their objective function using the $L2$-norm loss[29], [30], [31],[32] ,[33],[34],[35] ,[36]aiming to accurately and stably reconstruct the missing data in the non-standard tensor. However, the presence of outlier values significantly undermines the robustness of these models. How to effectively handle these outliers for robust and accurate prediction remains an open and urgent research problem. To improve the performance of tensor completion of traffic data containing various types of anomalies, this paper proposes a robust loss function. By setting a distance threshold between predicted and true values, the proposed loss assigns a fixed weight to samples within the threshold range, thereby maintaining the model's ability to learn from regular patterns while reducing its sensitivity to outliers.
Empirical results derived from two city road traffic average speed datasets demonstrate that TDWLFT outperforms existing traffic data recovery methods in terms of both prediction accuracy and computational efficiency for addressing missing data in the traffic data tensor.

The structure of this paper is outlined as follows: Section 2 introduces the preliminaries, Section 3 details the proposed methodology, Section 4 discusses the experimental results in depth, and Section 5 provides the conclusion.

## 2. Preliminaries

### 2.1. Problem of Data Recovery in ITS

In general, the data collected from ITS are often incomplete for a variety of reasons, such as sensor corruption, communication failure, unsatisfied sensor coverage, human errors, etc. The primary challenge in data recovery within ITS is to reconstruct the missing data based on the available observations. In the following section, we present a formalization of this issue.

**Definition 1**(*Problem of Data Recovery in ITS*): Let *I*, *J* and *K* be three large entity sets representing sensors, time intervals, and days, respectively. A third-order tensor $\mathbf{Y}^{|I|\times|J|\times|K|}$ is constructed to represent the traffic speeds observed across these dimensions, where each element $y_{ijk}$ denotes the traffic speed recorded by sensor $i \in I$ during time interval $j \in J$ on day $k \in K$. Let $\Lambda$ denote the set of know entries in **Y**, and $\Gamma$ represent the set of missing entries.

### 2.2 Latent Factorization of Tensors(LFT)

Let $\mathbf{Y}^{|I|\times|J|\times|K|}$ be a third-order tensor. An LFT [37] model aims to approximate **Y** by constructing a rank-*R* tensor $\hat{\mathbf{Y}}^{|I|\times|J|\times|K|} \in \mathbb{R}$[38].

**Definition 2**(*LFT Model*): Given a tensor **Y**, the LFT model approximates it using a sum of R rank-one tensors. Specifically, the approximation $\hat{\mathbf{Y}}$ is formulated as: $\hat{\mathbf{Y}} = \Sigma_{r=1}^{R} \mathbf{X}_r$, where each $\mathbf{X}_r^{|I|\times|J|\times|K|}$ denotes a rank-one tensor. To better understand this formulation, we next define what constitutes a rank-one tensor[39],[40],[35],[24].

**Definition 3**(*Rank-one Tensor*): A tensor $\mathbf{X}_r$ is said to be rank-one if it can be expressed as the outer product of three latent feature (LF) vectors: $\mathbf{X}_r = \boldsymbol{u}_r \circ \boldsymbol{s}_r \circ \boldsymbol{t}_r$, where $\boldsymbol{u}_r^{|I|}$, $\boldsymbol{s}_r^{|J|}$, and $\boldsymbol{t}_r^{|K|}$[41],[42],[43].

Thus, these LF vectors formulate three LF matrices $\mathbf{U}^{|I|\times R}$, $\mathbf{S}^{|J|\times R}$, and $\mathbf{T}^{|K|\times R}$. Each entry of $\mathbf{X}_r$, denoted $x_{ijk}^{(r)}$, can therefore be explicitly computed as:

$$x_{ijk}^{(r)} = u_{ir} s_{jr} t_{kr} . \quad (1)$$

Accordingly, the estimated value $\hat{y}_{ijk}$ within the reconstructed tensor $\hat{\mathbf{Y}}$ is defined as follows:

$$\hat{y}_{ijk} = \sum_{r=1}^{R} x_{ijk}^{(r)} = \sum_{r=1}^{R} u_{ir} s_{jr} t_{kr} . \quad (2)$$

In order to obtain the desired LF matrices U, S, and T, the reconstruction error between the original tensor and its low-rank approximation $\hat{\mathbf{Y}}$ is typically measured using the *L*2-norm[44], [45]. Given that the observed traffic tensor **Y** contains only a limited number of known entries, the loss function in LFT is formulated by focusing solely on the known subset $\Lambda$, in line with the principle of modeling based on density-oriented. Consequently, the optimization objective is defined using the *L*2-norm as follows[46],[47],[48],[37],[36],[49]:

$$\varepsilon = \frac{1}{2} \sum_{y_{ijk} \in \Lambda} \left( y_{ijk} - \hat{y}_{ijk} \right)^2 = \frac{1}{2} \sum_{y_{ijk} \in \Lambda} \left( y_{ijk} - \sum_{r=1}^{R} u_{ir} s_{jr} t_{kr} \right)^2 \quad (3)$$

Naturally, due to the uneven distribution of observed entries in **Y** and the model's high dependence on the initial values of U, S, and T, problem (3) becomes inherently ill-posed[27],[47]. To address this challenge and enhance the robustness and generalization capability of the model, it is essential to introduce Tikhonov regularization as a stabilizing mechanism, yielding

$$\varepsilon = \frac{1}{2} \sum_{y_{ijk} \in \Lambda} \left( y_{ijk} - \sum_{r=1}^{R} u_{ir} s_{jr} t_{kr} \right)^2 + \frac{\lambda}{2} \sum_{y_{ijk} \in \Lambda} \left( \sum_{r=1}^{R} \left( u_{ir}^2 + s_{jr}^2 + t_{kr}^2 \right) \right) \quad (4)$$

## 3. Methodology

### 3.1 Threshold Distance Weighted Loss Function

To achieve outlier resistance, this paper adopts the threshold distance weighted loss (TDW) to measure the distance between **Y** and $\hat{\mathbf{Y}}$. Given a threshold $\tau \in \mathbb{R}$, the threshold distance is defined as $|y_{ijk}-\tau|$, and the TDW can be expressed as follows:

$$\Delta_{ijk} = y_{ijk} - \sum_{r=1}^{R} u_{ir} s_{jr} t_{kr} \Rightarrow$$

$$\varepsilon = \begin{cases} \sum_{y_{ijk} \in \Lambda} \left( \Delta_{ijk} \right)^2, & |\Delta_{ijk}| \geq |y_{ijk} - \tau| \\ \sum_{y_{ijk} \in \Lambda} |y_{ijk} - \tau| |\Delta_{ijk}|, & |\Delta_{ijk}| < |y_{ijk} - \tau| . \end{cases} \quad (5)$$

where $\Delta_{ijk}$ denotes the prediction error between $y_{ijk}$ and $\hat{y}_{ijk}$. During the training process, the TDW loss applies differentiated weights to samples through the selection of the parameter $\tau$, effectively suppressing the impact of outliers on the model's convergence speed. Specifically, for observation points close to $\tau$, the prediction error is usually large, making the use of the mean squared error (MSE) more favourable for optimization. In contrast, for sample $y_{ijk}$ that deviate significantly from $\tau$, the weighted mean absolute error is applied with fixed weights to ensure their participation in training. For samples with a moderate distance from τ—i.e., those whose threshold distance and prediction error tend to fluctuate—adaptive weighting is employed based in their predictive performance. The position of the parameter *τ* reflects, to some extent, the model's area of focus within the data. The closer an observation value is to *τ*, the more likely the model is to optimize using MSE, enabling rapid learning and convergence for those samples. Therefore, the value of *τ* should be carefully chosen to avoid overemphasizing outliers, thereby enhancing robustness while preserving sensitivity to critical information. In this paper, the parameter *τ* is defined as the median value of the observed entries $y_{ijk}$.

*3.2 SGD-Based Learning Rules[50]*

By introducing a distance threshold $|y_i-\tau|$, the TDW assigns different optimization intensities to samples based on their prediction distances, thereby regulating the overall convergence speed of the model. An SGD algorithm[51],[52],[53],[52],[54],[55] is employed to solve (5) with respect to U, S, and T, as it is highly effective in optimizing the desired LFs

$$\begin{cases} u_{ir}^{n+1} \leftarrow u_{ir}^n - \eta\left(\partial \varepsilon_{ijk}^n / \partial u_{ir}^n\right) \\ s_{jr}^{n+1} \leftarrow s_{jr}^n - \eta\left(\partial \varepsilon_{ijk}^n / \partial s_{jr}^n\right) \\ t_{kr}^{n+1} \leftarrow t_{kr}^n - \eta\left(\partial \varepsilon_{ijk}^n / \partial t_{kr}^n\right) \end{cases} \quad (6)$$

where *η* denotes the learning rate. With (5) and (6), the following learning scheme is obtained:

$$\begin{cases} |\Delta_{ijk}| \geq |y_{ijk}-\tau|: \begin{cases} u_{ir}^{n+1} \leftarrow u_{ir}^n - \eta\left(-2\Delta_{ijk}s_{jr}^n t_{kr}^n + \lambda u_{ir}^n\right) \\ s_{jr}^{n+1} \leftarrow s_{jr}^n - \eta\left(-2\Delta_{ijk}u_{ir}^n t_{kr}^n + \lambda s_{jr}^n\right) \\ t_{kr}^{n+1} \leftarrow t_{kr}^n - \eta\left(-2\Delta_{ijk}u_{ir}^n s_{jr}^n + \lambda t_{kr}^n\right) \end{cases} \\ |\Delta_{ijk}| < |y_{ijk}-\tau|: \begin{cases} u_{ir}^{n+1} \leftarrow u_{ir}^n - \eta\left(-|y_{ijk}-\tau|\cdot sign(\Delta_{ijk})s_{jr}^n t_{kr}^n + \lambda u_{ir}^n\right) \\ s_{jr}^{n+1} \leftarrow s_{jr}^n - \eta\left(-|y_{ijk}-\tau|\cdot sign(\Delta_{ijk})u_{ir}^n t_{kr}^n + \lambda s_{jr}^n\right) \\ t_{kr}^{n+1} \leftarrow t_{kr}^n - \eta\left(-|y_{ijk}-\tau|\cdot sign(\Delta_{ijk})u_{ir}^n s_{jr}^n + \lambda t_{kr}^n\right) \end{cases} \end{cases}$$

(7)

where *n* and (*n*+1) denote the *n*th and the (*n*+1)-th update points of the LF matrices, *η* represents the learning rate, λ denotes the regularization coefficient respectively. With (7), the learning scheme is arrived.

## 4 Results

*4.1 General Settings*

**Datasets**: We conducted an empirical analysis using publicly available datasets of roadway speeds from two distinct cities, including Guangzhou and Seattle, their details are described as follows:
**D1: Guangzhou** Traffic Speed dataset [56]. This dataset consists of traffic speed information gathered from 214 sensors over a period of two-month (61 days, from August 1 to September 30, 2016) with 10-minute intervals in Guangzhou, China.
**D2: New York** Speed dataset. This data set is created by Uber movement project, which contains speed data gather from 135 sensors over 73 days (form October 1 to December 12, 2022) at 5-minute intervals in New York, USA.

Table 1 Dataset details

| No. | Dataset | Sensor Count | Time Slots | Day Count | Known entry Count |
|---|---|---|---|---|---|
| **D1** | Guangzhou | 214 | 144 | 61 | 1,855,589 |
| **D2** | New York | 135 | 288 | 73 | 2,193,015 |

To minimize bias, each dataset is randomly partitioned into three mutually exclusive subsets: a training set K, a validation set Ψ, and a testing set Ω, in accordance with a 7:1:2 ratio[57],[58]. The model is trained on K, validated using Ψ, and tested on Ω to obtain the final results. To account for variability introduced by random partitioning, this process is repeated 20 times[59],[60],[61],[62] producing multiple sets of experimental outcomes. The mean and standard deviation of the results are reported to assess the stability of the findings[63],[64],[65],[66].
For each model, the training process terminates under either of the following conditions 1) the number of iterations reaches 1,000 or 2) the reduction in validation error between two successive iterations falls below $10^{-5}$[67],[68].
**Model Settings:** To ensure reliable results, the following configurations are applied:
1) For each experiment conducted on the same dataset, the LF matrices are initialized with identical values to mitigate the impact of initialization bias.
2) The LF space dimension *R* is uniformly set to 20 for all models[69],[70]. This setting is chosen to strike a trade-off between computational efficiency and representation learning capacity, in alignment with the configuration used in [71].

**Evaluation Metric:** We evaluate the model's performance in tensor representation learning by measuring the reconstruction accuracy using two metrics: Root Mean Squared Error (RMSE) and Mean Absolute Error (MAE)[49],[72],[73],[74],[75]:

$$\text{RMSE} = \sqrt{\sum_{y_{ijk} \in \Psi} \left( y_{ijk} - \hat{y}_{ijk} \right)^2 \Big/ |\Omega|}$$

$$\text{MAE} = \sum_{y_{ijk} \in \Psi} \left| y_{ijk} - \hat{y}_{ijk} \right| \Big/ |\Omega|$$

Lower RMSE and MAE values indicate higher accuracy in predicting missing values within a tensor, reflecting a stronger representation learning ability[76],[77, 78],[79].

*4.2 Comparison with State-of-the-art Models*

We evaluate TDWLFT by benchmarking its estimation precision and computational speed against multiple state-of-the-art models.

**M1**: The proposed model in this study.

**M2**: It employs a process that is conceptually the reverse of CP decomposition [80], where the values of all factor matrices are updated by alternating least squares or by gradient descent algorithms.

**M3**: A low-rank autoregressive tensor completion framework [81], which imposes an autoregressive model on each time series with coefficients and employs an alternating minimization scheme to estimate the low-rank tensor.

**M4:** Reference [82] proposes an efficient nonparametric tensor decomposition method specifically designed for binary and count data. It employs a nonparametric Gaussian process and Pólya-Gamma augmentation to establish conjugate models.

Table 2 presents the RMSE and MAE of M1-4 on D1-2, while Fig. 1 visualizes these results. Table 3 provides a summary of their computational efficiency. Based on the aforementioned experimental results, the following conclusions can be drawn.

Table 2 Prediction accuracy (RMSE and MAE) of all tested models on D1-2

| Dataset | | M1 | M2 | M3 | M4 |
|---|---|---|---|---|---|
| D1 | RMSE | 4.6966 | 4.7605 | 4.7971 | 5.5678 |
| | MAE | 3.1622 | 3.2274 | 3.2408 | 3.7796 |
| D2 | RMSE | 9.3807 | 9.5786 | 9.4870 | 10.8469 |
| | MAE | 6.0269 | 6.3705 | 7.2443 | 7.5780 |

Table 3 Time costs of all tested models on D1-D2

| Dataset | | M1 | M2 | M3 | M4 |
|---|---|---|---|---|---|
| D1 | Time-RMSE | 150 | 300 | 502 | 473 |
| | Time-MAE | 150 | 450 | 502 | 473 |
| D2 | Time-RMSE | 175 | 1025 | 912 | 635 |
| | Time-MAE | 175 | 1025 | 924 | 635 |

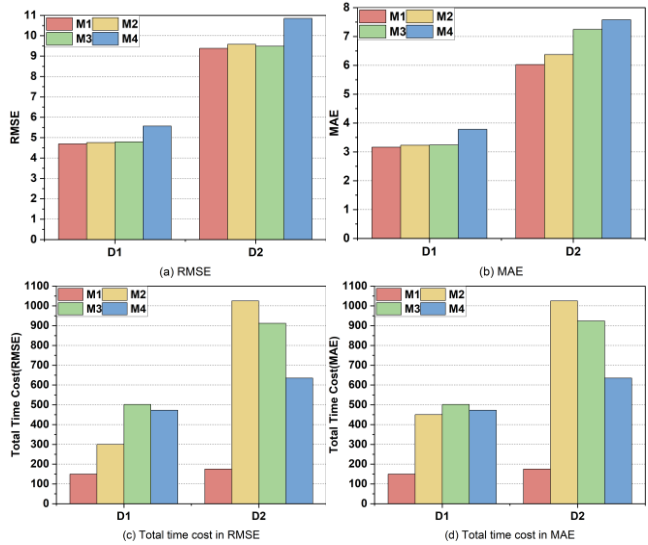

Fig. 1 Performance comparison of M1-M4 on D1-D2

1) **M1, i.e., a proposed TDWLFT model, outperforms its peers in terms of prediction accuracy for missing data of a nonstandard tensor.** As shown in Figs. 1(a)-(b) and Table 2, M1 achieves the lowest RMSE and MAE on D1-2. For instance, on D1, M1 obtains lowest RMSE at 4.6966, which is about 1.34% lower than 4.7605 by M2 (RMSE$_{M2}$-RMSE$_{M1}$/RMSE$_{M2}$), 2.09% lower than 4.7971 by M3, 15.64% lower than 5.5678 by M4. Considering MAE, on D1, the output by M1-4 are 3.1622, 3.2274, 3.2408, 3.7796, respectively. Hence, M1's MAE is also evidently lower than that of its peers. Similar results can be drawn on D2, as summarized in Table 2 and Fig. 1(a)-(b).

2) **M1's computational efficiency is remarkably efficient in comparison to its peers.** As shown in Table 3 and Figs. 1(c)-(d), on D1, M1's total time cost for RMSE and MAE is 150 seconds, which are the lowest among all models. By comparison, M2's total time cost for achieving the lowest RMSE and MAE is approximately 300 seconds and 450 seconds, which is considerably higher than M1's total time cost. Likewise, on D2, the time cost of M3-M4 in RMSE and MAE is also substantially higher than that of M1. A similar trend can be observed on D2, as illustrated in Table 3 and Figs. 1(c)-(d).

## 5 Conclusion

This paper proposes a threshold distance weighted loss-incorporated Latent Factorization of Tensors (TDWLFT) model, which is designed to effectively reconstruct incomplete traffic data within urban road networks. The TDWLFT model leverages the threshold distance weighted loss to construct a robust objective function[83]. Through extensive experimental validation, we demonstrate that the TDWLFT model achieves superior performance compared to existing state-of-the-art models, significantly reducing time costs while enhancing recovery accuracy in spatiotemporal traffic data reconstruction tasks. Future research should explore the utilization of adaptive regularization term adjustment to enhance model robustness and adoption of advanced objective function to improve data recovery performance [84]. These challenges will be systematically explored in future research efforts.